\documentclass[runningheads]{llncs}
\usepackage{amssymb}
\usepackage{multirow,setspace,verbatim,amsfonts,amsmath,amsbsy,epsfig,url}
\usepackage{gensymb}
\usepackage{booktabs}
\usepackage{cite}
\usepackage{makecell}
\usepackage{epstopdf}
\usepackage{array,multirow}
\newcolumntype{P}[1]{>{\centering\arraybackslash}p{#1}}
\newcolumntype{M}[1]{>{\centering\arraybackslash}m{#1}}
\usepackage{algorithm}
\usepackage[noend]{algpseudocode}
\usepackage{tikz}
\usepackage{mathrsfs} 
\usepackage[algo2e]{algorithm2e} 
\usepackage{array}
\usepackage{color}
\usepackage{graphicx}
\begin{document}
\title{Unsupervised Deformable Image Registration Using Cycle-Consistent CNN\thanks{This work was supported by the Industrial Strategic technology development program (10072064, Development of Novel Artificial Intelligence Technologies To Assist Imaging Diagnosis of Pulmonary, Hepatic, and Cardiac Diseases and Their Integration into Commercial Clinical PACS Platforms) funded by the Ministry of Trade Industry and Energy (MI, Korea).}}
%
%\titlerunning{Abbreviated paper title}
% If the paper title is too long for the running head, you can set
% an abbreviated paper title here
%
\author{Boah Kim\inst{1}\orcidID{0000-0001-6178-9357} \and %
Jieun Kim\inst{2} \and
June-Goo Lee\inst{2} \and \\
Dong Hwan Kim \inst{2} \and
Seong Ho Park\inst{2} \and
Jong Chul Ye\inst{1}\orcidID{0000-0001-9763-9609}} %

% index{Kim, Boah} 
% index{Kim, Jieun} 
% index{Lee, June-Goo} 
% index{Kim, Dong Hwan} 
% index{Park, Seong Ho} 
% index{Ye, Jong Chul} 

\authorrunning{B. Kim et al.}
%% First names are abbreviated in the running head.
%% If there are more than two authors, 'et al.' is used.
%%
\institute{Korea Advanced Institute of Science and Technology, Daejeon, South Korea \\
\email{\{boahkim, jong.ye\}@kaist.ac.kr}\\ \and
University of Ulsan College of Medicine, Asan Medical Center, Seoul, South Korea} 

\maketitle              % typeset the header of the contribution

\begin{abstract}
Medical image registration is one of the key processing steps for  biomedical image analysis such as cancer diagnosis. Recently, deep learning based supervised and unsupervised image registration methods have been extensively studied due to its excellent performance in spite of ultra-fast computational time compared to the classical approaches. In this paper, we present a novel unsupervised medical image registration method that trains deep neural network for deformable registration of 3D volumes using a cycle-consistency. Thanks to the cycle consistency, the proposed deep neural networks can take diverse pair of image data with severe deformation for accurate registration. Experimental results using multiphase liver CT images demonstrate that our method provides very precise 3D image registration within a few seconds, resulting in more accurate cancer size estimation.
%153words 
\end{abstract}
\section{Introduction}
Radiologists often diagnose the progress of disease by comparing medical images at different temporal phases. In case of diagnosis of liver tumor such as hepatocellular carcinoma (HCC), the contrast of normal liver tissue and tumor region in contrast enhanced CT (CECT) distinctly varies before and after the infection of contrast agent. This provides radiologists an important clue to diagnose cancers and plan surgery or radiation therapy \cite{kim2011assessment}. However, liver images taken at different phases are usually different in their shape due to disease progress, breathing, patient motion, etc, so image registration is important to improve accuracy of dynamic studies.

Classical image registration methods \cite{thirion1998image, christensen2001consistent} are usually implemented in a variational framework that solves an energy minimization problem over the space of deformations. Since the diffeomorphic image registration ensures the preservation of topology and one-to-one mapping between the source and target images, the algorithmic extensions to large deformation such as LDDMM \cite{beg2005computing} and SyN \cite{avants2008symmetric} have been applied to various image registration studies. However, these approaches usually require substantial time and extensive computation. 

To address this issue, recent image registration techniques are often based on deep neural networks that instantaneously generate deformation fields. In supervised learning approaches \cite{yang2017quicksilver, zhu2017unpaired}, the ground-truths of deformation fields are required for training neural networks, which are typically generated by the traditional registration method. However, the performance of these existing supervised methods depends on the quality of the ground-truth registration fields, or they do not explicitly enforce the consistency criterion to uniquely describe the correspondences between two images. 

In order to overcome the aforementioned limitations and provide topology-preserving guarantee, many unsupervised learning methods have been developed in these days. Balakrishnan et al. \cite{balakrishnan2018unsupervised} propose 3D medical image registration algorithm using a spatial transform network. Zhang \cite{zhang2018inverse} presents a CNN framework that enforces an inverse-consistent constraint for the deformation fields. However, for the registration of large deformable volumes such as livers, the existing unsupervised learning methods often result in inaccurate registration due to the potential for the degeneracy of the mapping. Although Dalca et el. \cite{dalca2018unsupervised} tried to address this problem by incorporating a diffeomorphic integration layer, we found that its application of the liver registration is still limited.

In this paper, we present a novel unsupervised registration method using convolutional neural networks (CNN) with cycle-consistency \cite{zhu2017unpaired}. We show that the cyclic constraint can be adopted for the image registration case naturally, and this cycle consistency improves topology preservation in generating fewer folding problems. Also, our network is trained with diverse source and target images in multiphase CECT acquisition so that a single neural network of our method provides deformable registration between every pairs once the network is trained. Experimental results demonstrate that the proposed method performs accurate 3D image registration for any pair of images within a few seconds in the challenging problem of 3D liver registration in multiphase CECT.

%----------------------------------------------------------------------------------------------------------
%----------------------------------------------------------------------------------------------------------
\section{Proposed Method}

\begin{figure}[t!]
\centering
\includegraphics[width=11cm]{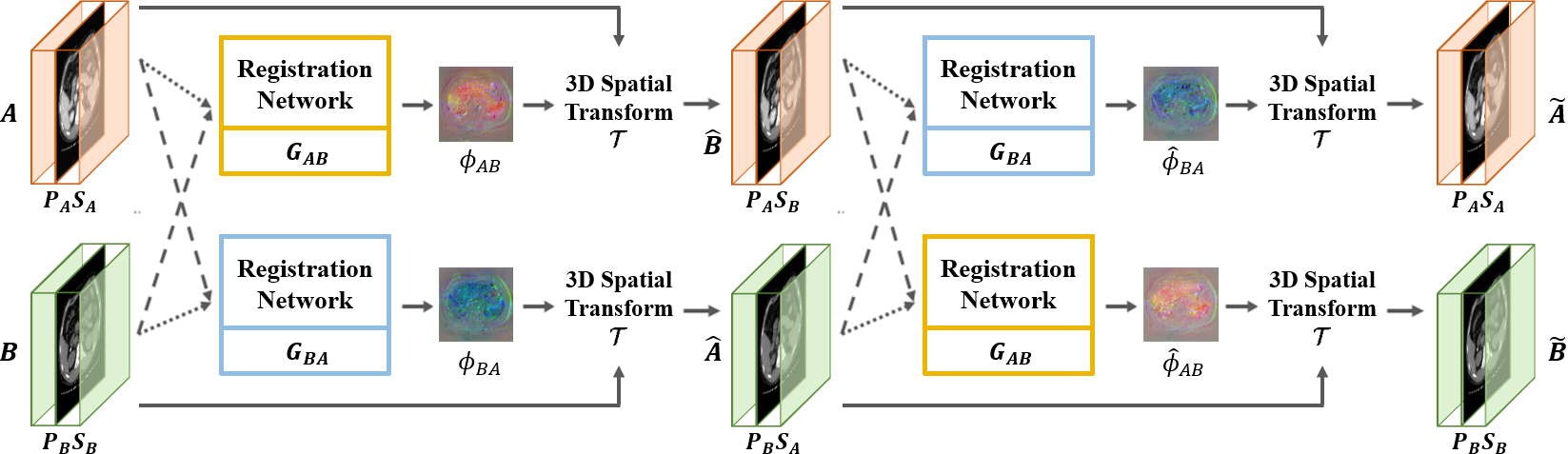}
\caption{The overall framework of the proposed method for image registration. The input images in different phases are denoted as $A$ and $B$, and their phase and shape are denoted as $P$ and $S$, respectively. The short-dashed line indicates the floating image and the long-dashed line denotes the fixed image.}
\label{fig:pipeline}
\end{figure}

The overall framework of our method is illustrated in Figure \ref{fig:pipeline}. For the input images, $A$ and $B$, in different phases, we define two registration networks as $G_{AB}:(A, B) \rightarrow \phi_{AB}$ and $G_{BA}:(B, A) \rightarrow \phi_{BA}$, where $\phi_{AB}$ (resp. $\phi_{BA}$) denotes the 3-D deformation field from $A$ to $B$ (resp. $B$ to $A$). We use a 3D spatial transformation layer $\mathcal{T}$ in the networks to warp the moving image by the
estimated deformation field, so that the registration networks are trained to minimize the dissimilarity between the deformed moving source image and fixed target image. Accordingly, once a pair of images are given to the registration networks, the moving images are deformed into the fixed images.

To guarantee the topology preservation between the deformed and fixed images, we here adopt the cycle consistency constraint between the original moving image and its re-deformed image. That is, the deformed volumes are given as the inputs to the networks again by switching their order to impose the cycle consistency. This constraint ensures that the shape of deformed images successively returns to the original shape. 

%-------------------------------------
%-------------------------------------

\subsection{Loss Function}
\begin{figure}[t!]
\centering
\includegraphics[width=10cm]{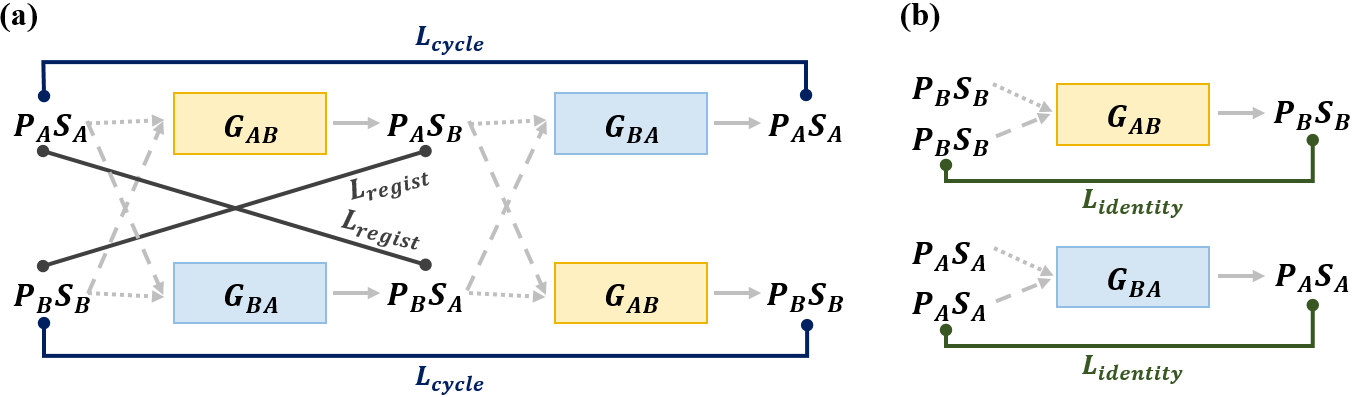}
\caption{The diagram of loss function structure in our proposed method. The short- and long-dashed lines are for floating image and fixed image, respectively.}
\label{fig:total_loss}
\end{figure}

We train the networks by solving the following loss function:
\begin{align}
\mathcal{L} 
= \mathcal{L}&_{regist}^{AB} + \mathcal{L}_{regist}^{BA}
 + \alpha \mathcal{L}_{cycle}
+ \beta \mathcal{L}_{identity}, 
\label{eq:loss}
\end{align}
where $\mathcal{L}_{regist}$, $\mathcal{L}_{cycle}$, and $\mathcal{L}_{identity}$ are registration loss, cycle loss, and identity loss, respectively (see Fig. \ref{fig:total_loss}), and $\alpha$ and $\beta$ are hyper-parameters. Based on this loss function, our method is trained in an unsupervised manner without ground-truth deformation fields. 

%-------------------------------------
\noindent \textbf{Registration Loss.}
The registration loss function is based on the energy function of classical variational image registration. For example, the energy function for the registration of floating image $A$ to the target volume $B$ is composed of two terms:
\begin{align}
%\mathcal{L}
 \mathcal{L}&_{regist}^{AB}= \mathcal{L}_{sim}\left(\mathcal{T}(A, \phi), B\right) + \mathcal{L}_{reg}({\phi}),
\label{eq:optimization problem}
\end{align}
where $A$ is the moving image, and $B$ is the fixed image. $\mathcal{L}_{sim}$ computes image dissimilarity between the deformed image by the estimated deformation field $\phi$ and the fixed image, and $\mathcal{L}_{reg}$ evaluates the smoothness of the deformation field. Here, $\mathcal{T}$ denotes the 3D spatial transformation function.
In particular, we employ the cross-correlation as the similarity function to deal with the contrast change during
CECT exam, and the L2-loss for regularization function. Accordingly, our registration loss function can be written as:
\begin{align}
\mathcal{L}_{regist}^{AB} = -(\mathcal{T}(A, \phi_{AB}) \otimes B ) +\lambda ||\phi_{AB}||_2,
\label{eqn:regist_loss}
\end{align}
where $\otimes$ denotes the cross-correlation defined by
\begin{align}
& x\otimes y = \frac{ |\langle x-\bar x, y-\bar y \rangle|^2}{ \| x-\bar x\| \| y-\bar y\|},
\label{eqn:cross-correlation}
\end{align}
where $\bar x$ and $\bar y$ denote the mean value of $x$ and $y$, respectively.

%-------------------------------------
\noindent \textbf{Cycle Loss.}
The cycle consistency condition is implemented by minimizing the loss function as shown in Fig. \ref{fig:total_loss}(a).
Since an image $A$ is first deformed to an image $\hat B$, which is then deformed again by the other network to generate image $\tilde A$, the cyclic consistency imposes $A\simeq \tilde A$. Similarly, an image $B$ should be successively deformed by the two networks to generate image $\tilde B$. Then, the cyclic consistency is to impose that $B\simeq \tilde B$.

As shown in Fig. \ref{fig:total_loss}(a), since the network in our registration receives both the moving image and the fixed image, the implementation of the cycle consistency loss should be given by as the vector-form of the cycle consistency condition:
\begin{eqnarray}
\begin{bmatrix} A\\ B\end{bmatrix}
\simeq \begin{bmatrix} 
 \mathcal{T}(\hat B, \hat \phi_{BA})\\  \mathcal{T}(\hat A, \hat \phi_{AB}) 
 \end{bmatrix} 
 =
 \begin{bmatrix} 
 \mathcal{T}\left(\mathcal{T}(A, \phi_{AB}), \hat \phi_{BA}\right)\\  \mathcal{T}\left( \mathcal{T}(B, \phi_{BA}), \hat \phi_{AB}\right)
 \end{bmatrix}
\end{eqnarray}
where 
$ (\hat B, \hat A):= \left(\mathcal{T}(A, \phi_{AB}), \mathcal{T}(B, \phi_{BA})\right).$
Thus, the cycle loss is computed by:
\begin{align}
\mathcal{L}_{cycle} = \left\|\begin{bmatrix} 
 \mathcal{T}(\hat B, \hat \phi_{BA})\\  \mathcal{T}(\hat A, \hat \phi_{AB}) 
 \end{bmatrix} 
-\begin{bmatrix} A\\ B\end{bmatrix} \right\|_1,
\label{eqn:cycle_loss}
\end{align}
where $||\cdot||_1$ denotes the $l_1$-norm.

%-------------------------------------
\noindent \textbf{Identity Loss.}
Another important consideration for the design of loss function is that the network should not change the stationary regions of the body, i.e.
the stationary regions should be the fixed points of the network. As shown in Fig. \ref{fig:total_loss}(b), this constraint can be implemented by imposing that the input image should not be changed when the identical images are used as the floating and reference volumes. More specifically, we use the following identity loss:
\begin{align}
\mathcal{L}_{identity}  = -(\mathcal{T}(A, G_{AB}(A,A)) \otimes A) - (\mathcal{T}(B, G_{BA}(B,B)) \otimes B ).
\label{eqn:identity_loss}
\end{align}
By minimizing this identity loss \eqref{eqn:identity_loss}, the cross-correlation between the deformed image and the fixed image can be maximized. Thus, the identity loss guides the stability of the deformable field estimation in stationary regions.

%-----------------------------------------------------------------------------------------
\subsection{Network Architecture and 3D Spatial Transformation Layer}
To generate a displacement vector field in width-, height-, and depth direction, we adopt VoxelMorph-1 \cite{balakrishnan2018unsupervised} as our baseline network. Note that our model without both the cycle and identity loss is equivalent to VoxelMorph-1. This 3D network consists of encoder, decoder and their skip connections similar to U-Net \cite{ronneberger2015u}. %The encoder typically extracts a hierarchy of image feature maps from low to high complexity, while the decoder transforms the features and reconstructs the output from low to high resolution. The encoder-decoder skip connections play a key role in the decoder to use high resolution features from the encoder as an extra inputs.

The 3D spatial transformation layer \cite{jaderberg2015spatial} is to deform the moving volume with the deformation field $\phi$. We use the spatial transformation function $\mathcal{T}$ with tri-linear interpolation for warping the image $A$ by $\phi$, which can be written as:
\begin{align}
\mathcal{T}(A, \phi) = \sum\nolimits_{y \in \mathcal{N}(x+\phi(x))}^{ }A(y)\prod\nolimits_{d \in \{i,j,k\}}^{ }(1-|x_d + \phi(x_d)-y_d|),
\label{eq:st}
\end{align}
where $x$ indicates the voxel index, $\mathcal{N}(x+\phi(x))$ denotes the 8-voxel cubic neighborhood around $x+\phi(x)$, and $d$ is three directions in 3D image space.

%-----------------------------------------------------------------------------------------
%-----------------------------------------------------------------------------------------
\section{Experiments}
To verify the performance of our method, we conducted liver registration from multiphase CT images. The dataset was collected from liver cancer (HCC) patients at Asan Medical Center, Seoul, South Korea. Each scans has pathologically proven hepatic nodules and four-phase liver CT (unenhanced, arterial, portal, and 180-s delayed phases). The slice thickness was 5mm. We did not perform pre-processing such as affine transformation except for matching the number of slices for the moving and fixed images. Here, we extracted slices only including liver by a pre-trained liver segmentation network and performed zero-padding to the above and below volumes based on the center of mass of liver.

We used 555 scans for training and 50 scans for testing. For the network training, we stacked two volumes with different phases as the input. We normalized the input intensity with the maximum value of each volume. Also, we randomly down-sampled the training data from $512\times 512\times depth$ to $128\times 128\times depth$ to fit in the GPU memory, while we evaluated the test data with original size of $512\times 512\times depth$, where $depth$ is different for each pair of input. For data augmentation, we adopted random horizontal/vertical flipping and rotation with 90 degree for each pair of training volume. Our proposed method was implemented with pyTorch library. We applied Adam with momentum optimization algorithm to train the models with a learning rate of $0.0001$, and set the batch size 1. The model was trained for 50 epochs using a NVIDIA GeForce GTX 1080 Ti GPU.

\begin{figure}[b!]
\centering
\includegraphics[width=12cm]{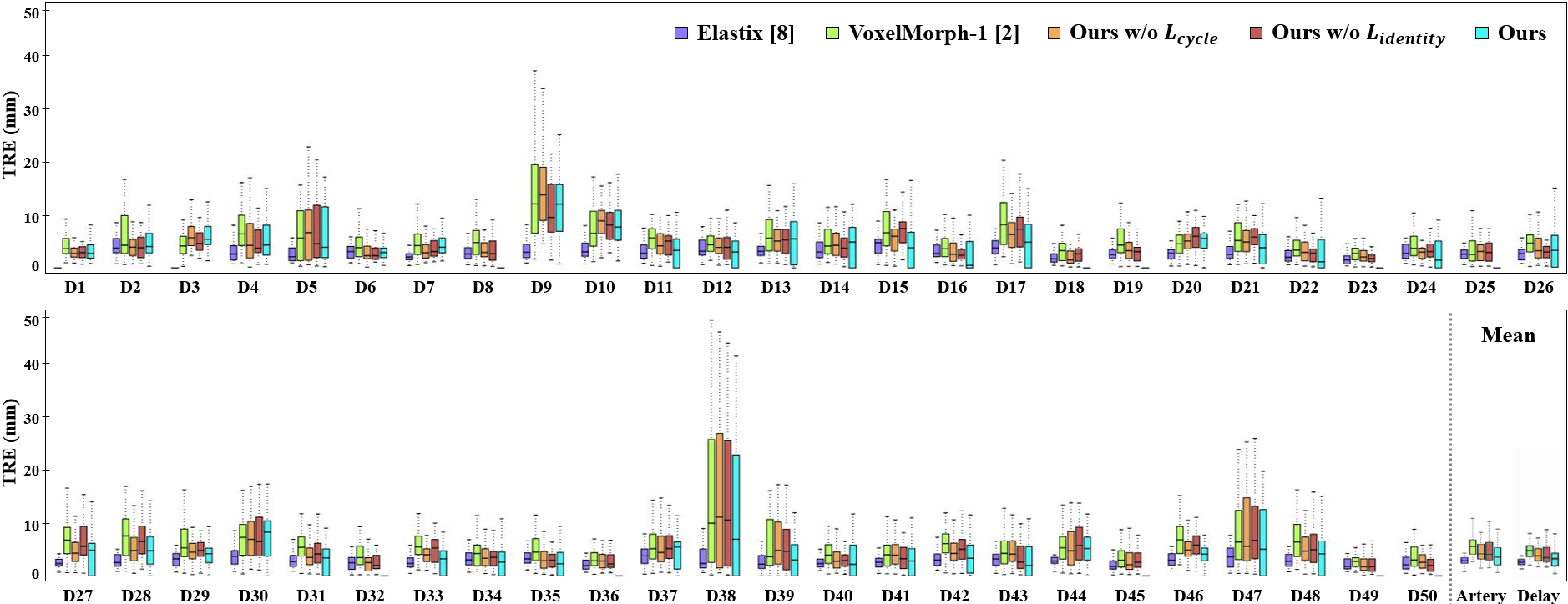}
\caption{Results of the target registration errors (TRE) of all 20 anatomical points in the deformed arterial and delayed images of 50 test data. Mean graph represents the mean TRE of the points for all subjects. D{\#} in the x-axis indicates the patient number.}
\label{fig:results_graph}
\end{figure}

\begin{table}[b!]
  \caption{Tumor size differences, TRE values ($mm$) between the deformed arterial/delayed images and the fixed portal image, and their average time (min) to be deformed on test set.}
  \label{table:size}
  \centering
  \begin{tabular}{M{2.6cm}|M{1.1cm}|M{1.1cm}|M{1cm}|M{1cm}|M{1.1cm}|M{1.1cm}|M{1cm}|M{1cm}}
  \hline
  \multirow{3}{*} {\textbf{Method}} & \multicolumn{4}{c|}{\textbf{Arterial $\rightarrow$ Portal}} & \multicolumn{4}{c}{\textbf{Delayed $\rightarrow$ Portal}}\\ 
  \cline{2-9}
  & \multicolumn{2}{c|}{tumor size} & \multirow{2}{*}{TRE}& \multirow{2}{*}{time}  & \multicolumn{2}{c|}{tumor size} & \multirow{2}{*}{TRE} & \multirow{2}{*}{time} \\
  \cline{2-3} \cline{6-7}
  & {major} & {minor} & & &{major} & {minor} & &\\
  \hline
   \multicolumn{1}{l|}{Elastix \cite{klein2010elastix}} & 0.98 & \textbf{0.61} & \textbf{3.26} & 19.64 & 0.91 & 0.58 & \textbf{2.96} & 19.64 \\ 
  \multicolumn{1}{l|}{VoxelMorph \cite{balakrishnan2018unsupervised}} & \textbf{0.79} & 1.64 & 6.67 & 0.18 & 0.61 & 0.87 & 5.35 & 0.20 \\
   \multicolumn{1}{l|}{\textbf{Ours}}  & 0.89 & 1.16 & 4.91 & 0.22 & \textbf{0.59} & \textbf{0.43} & 3.76 & 0.20\\
   \hline
  \end{tabular}  
\end{table}

\subsection{Registration Results}
We evaluated the registration performance using the target registration error (TRE) based on the 20 anatomical points in the liver and adjacent organs at the portal phases, which are marked by radiologists. Also, we measured the tumor size that verifies the registration performance in the view point of tumor diagnosis. We compared our method to Elastix \cite{klein2010elastix} that is known to have its state-of-the-art performance among the classical approaches. We also compared with VoxelMorph-1 \cite{balakrishnan2018unsupervised}. Additionally, we performed ablation studies by excluding the cycle loss or identity loss. Apart from the loss, different ablated networks were subjected to the same training procedure for fair comparison. 

Fig. \ref{fig:results_graph} shows the registration performance. We visualize the TRE values of the deformed arterial and delayed images with respect to each test data, and also show the average TRE values of all subjects with respect to the deformed arterial and delayed images into the fixed portal image. 
We can observe that the proposed method achieves significant improvement compared to VoxelMorph-1, while the error of the proposed method is slightly higher than Elastix. Also, in Table \ref{table:size}, we can confirm that the tumor size of deformed images from our proposed method is the most accurate for the case of delay to portal registration, and comparable in arterial to portal registration. 

To demonstrate the effect of the cycle consistency, we also computed the percentage of voxels with a non-positive Jacobian determinant on the deformation fields and the normalized mean square error (NMSE) between the original moving image and re-deformed image. As shown in Table \ref{table:jacobian}, we confirm that the proposed method is less prone to folding problem and improves topological preservation for liver registration.

\begin{figure}[t!]
\centering
\includegraphics[width=12cm]{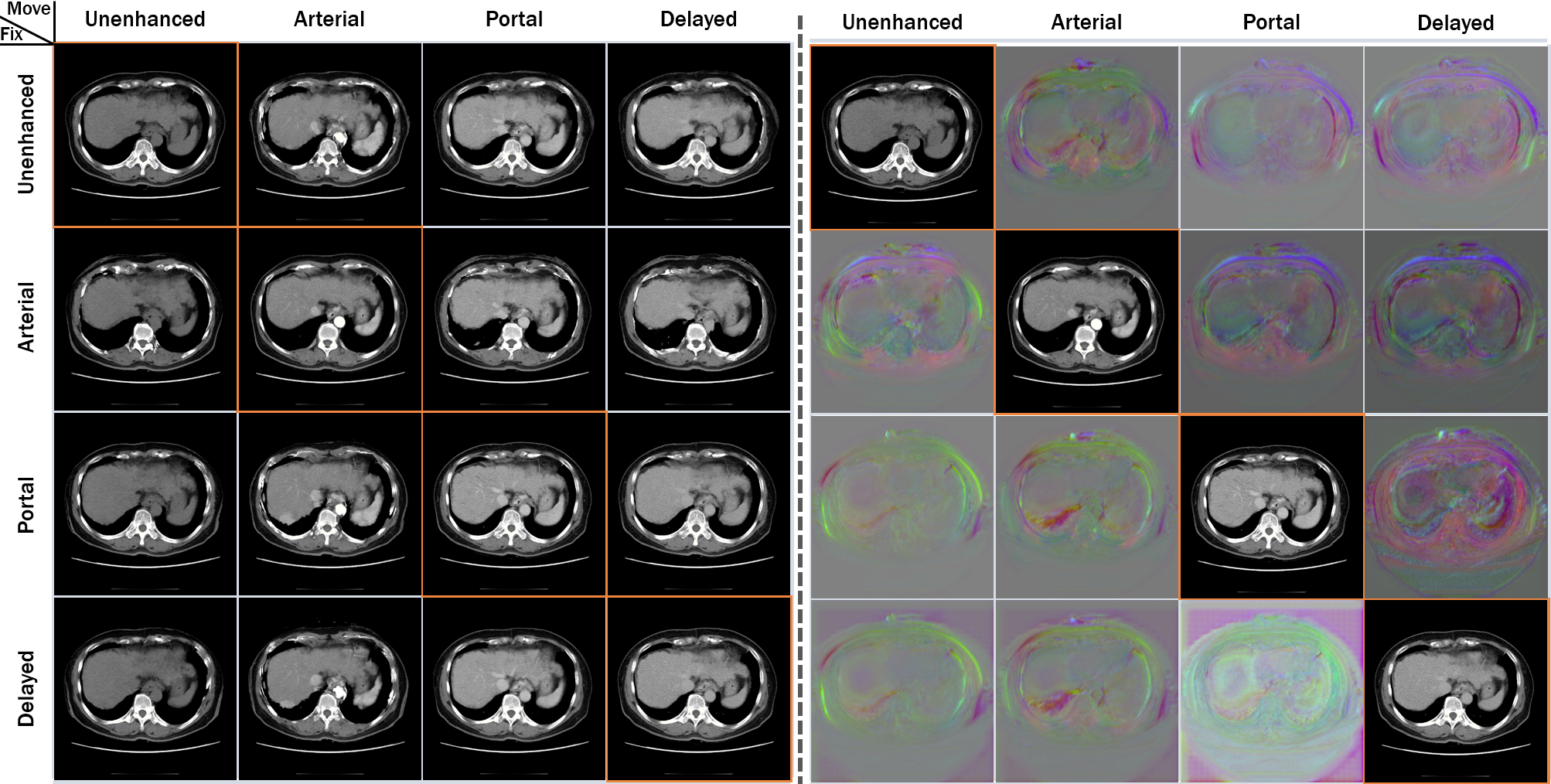}
\caption{Results of multiphase liver CT registration (Left) and their deformation fields (Right). The diagonal images with red-box are original images, which are deformed to other phases as indicated by each row. Specifically, the $(i,j), i\neq j$ element of the figure represents the deformed image to the $i$-th phase from the $j$-th phase original image.}
\label{fig:results_qual2}
\end{figure}

Fig. \ref{fig:results_qual2} illustrates an example of registration results that deforms the multiphase 3D images with the four distinct phases. Moreover, we calculated the time required for the proposed method to deform one image into the fixed image (see Table \ref{table:size}). Specifically, the conventional Elastix takes approximately 19.6 minutes for the image registration, while the proposed method takes only 10 seconds. 

\begin{table}[t!]
  \caption{Percentage of voxels with a non-positive Jacobian determinant and normalized mean square error (NMSE) on test set. }
  \label{table:jacobian}
  \centering
\begin{tabular}{M{3cm}|M{2.7cm}|M{1.5cm}|M{2.7cm}|M{1.5cm}}
  \hline
  \multirow{2}{*} {\textbf{Method}} & \multicolumn{2}{c|}{\textbf{Arterial $\rightarrow$ Portal}} & \multicolumn{2}{c}{\textbf{Delayed $\rightarrow$ Portal}}\\ 
  \cline{2-5}
  & \% of det$(J_\phi)\leq 0$ & NMSE & \% of det$(J_\phi)\leq 0$ & NMSE    \\
  \hline
  \multicolumn{1}{l|}{VoxelMorph \cite{balakrishnan2018unsupervised}} & {0.0327} & 0.0278 & 0.0311 & 0.0213\\
 \multicolumn{1}{l|}{Ours w/o $\mathcal{L}_{cycle}$  }& {0.0270} & 0.0279 & 0.0284 & 0.0214 \\
   \multicolumn{1}{l|}{Ours w/o $\mathcal{L}_{identity}$ } & {0.0218} & 0.0279 & 0.0205 & 0.0208 \\
   \multicolumn{1}{l|}{\textbf{Ours}}  & \textbf{0.0175} & \textbf{0.0277} & \textbf{0.0181} & \textbf{0.0199}\\
   \hline
  \end{tabular}  
\end{table}

%------------------------------------------------------------------------------------------------------------------
%------------------------------------------------------------------------------------------------------------------
\section{Conclusion}
We presented an unsupervised image registration method using a cycle consistent convolutional neural network. Using two registration networks, our proposed method is trained to satisfy the cycle consistency that imposes inverse consistency between a pair of images. However, once the networks are trained, a single network can provide accurate 3D image registration with any pair of new data, so the computational complexity is same as VoxelMorph-1. Our liver registration results demonstrated that the proposed method works well for any image pairs with different contrast.

%\bibliographystyle{splncs04}
%\bibliography{2019MICCAI_KBA_ref}
%
%% ---- Bibliography ----
%%
%% BibTeX users should specify bibliography style 'splncs04'.
%% References will then be sorted and formatted in the correct style.
%%

\end{document}